\documentclass[12pt]{article} 
\usepackage{newtxtext,newtxmath, amsmath, amsfonts, bm,mathtools}

\usepackage{paralist, url, natbib, parskip}
\usepackage{adjustbox, todonotes} 
\usepackage{graphicx, subfig}
\usepackage{rotating, booktabs, multirow}
\usepackage[makeroom]{cancel}
\usepackage[section]{placeins}
\usepackage[linesnumbered,lined,boxed,commentsnumbered]{algorithm2e}
\usepackage{xcolor}
\usepackage{placeins}
\usepackage{arydshln}
\usepackage{pgf,tikz,pgfplots}
\pgfplotsset{compat=1.15}
\usepackage{mathrsfs}
\usepackage{arydshln}
\usetikzlibrary{arrows}
\usetikzlibrary{decorations.markings}
\usetikzlibrary{positioning, shadows}
\graphicspath{{Figures/}}

\usepackage[margin=3cm]{geometry}
\usepackage[margin=2cm]{caption}
\usepackage{titling}
\usepackage{setspace}
\DeclareCaptionStyle{italic}[justification=centering]{labelfont={bf},textfont={it},labelsep=colon}
\SetKwInOut{Parameter}{parameter}
\mathtoolsset{showonlyrefs}
\usepackage{mathtools}
\usepackage{todonotes}
\usepackage{array}
\newcolumntype{R}[1]{>{\raggedleft\let\newline\\\arraybackslash\hspace{0pt}}m{#1}}
\newcolumntype{L}[1]{>{\raggedright\let\newline\\\arraybackslash\hspace{0pt}}m{#1}}
\newcolumntype{C}[1]{>{\centering\let\newline\\\arraybackslash\hspace{0pt}}m{#1}}

\usepackage{amsthm}
\usepackage{bm}
\usepackage{enumitem}

\usepackage[capitalize,noabbrev]{cleveref}
\theoremstyle{plain}
\newtheorem{theorem}{Theorem}[section]

\theoremstyle{definition}
\newtheorem{definition}[theorem]{Definition}

\theoremstyle{remark}

\newcommand{\dens}{\text{dens}}

\mathtoolsset{showonlyrefs=TRUE}

\title{Forecasting the structure of dynamic graphs}
\author{Sevvandi Kandanaarachchi, Ziqi Xu, Stefan Westerlund}
\date{}

\begin{document}

\maketitle

\begin{abstract}
Many aspects of graphs have been studied in depth. However, forecasting the structure of a graph at future time steps incorporating unseen, new nodes and edges has not gained much attention. In this paper, we present such an approach. Using a time series of graphs, we forecast graphs at future time steps.  We use time series forecasting methods to predict the node degree at future time points and combine these forecasts with flux balance analysis -- a linear programming method used in biochemistry -- to obtain the structure of future graphs. We evaluate this approach using synthetic and real-world datasets and demonstrate its utility and applicability. 

\end{abstract}


\section{Introduction}\label{sec:introduction}
We are surrounded by networks in different mediums and scales. From a biological perspective protein-protein interaction networks mediate cellular responses much needed for our biological function; from an infrastructure viewpoint, road, rail and airline  networks govern a significant portion of our time spent on commuting, and from an online aspect,  social media networks have been part of our lives since the latter part of the twentieth century. There is extensive research conducted on different aspects of networks/graphs including link prediction, anomaly detection, time series forecasting,  inference and embeddings to name a few. In this paper, we focus on forecasting the structure of dynamic, undirected, unweighted graphs at future time steps including new nodes and edges.


Dynamic graphs, such as those encountered in social networks change over time  with the addition and deletion of nodes and edges. When people join a social network new nodes are added, and when they leave nodes should ideally be deleted. Similarly, networks grow in epidemiological settings, where human movement and proximity play a big role. A somewhat different example is a home electricity network, where different appliances can plug in to different power sockets at different times for different durations. If each type of appliance is modelled as a separate class of nodes then we can model the  network as a heterogenous graph. Nodes such as phones move over the network when appliances connect to different power outlets at different times. Going beyond the home electricity network, this pattern is seen when electric cars charge at different locations at different time points. These are examples of dynamic graph structures.  In contrast, the structure of static graphs remains constant. For example, while the traffic changes over time in road networks, the underlying graph structure remains the same apart from incidents of road closures. While static graphs have been studied for longer, there is a lot of interest in dynamic graphs. 


Generally dynamic graphs are represented in two different formats: discrete-time dynamic graphs (DTDG) and continuous-time dynamic graphs (CTDG) \citep{Kazemi2020}. Discrete-time dynamic graphs consider a sequence of graphs at regular time intervals, for example, daily graphs on social network activity.  Thus, the data comes in the form of a time series of graphs. In contrast, continuous-time dynamic graphs (CTDG) are more event-driven, where each event can be the formation of an edge, the addition of a new node or similar activity. Timestamped edgelists are generally used in the continuous setting. Our focus on discrete-time dynamic graphs (DTDG) and as such we work with time series of graphs.

Of the diverse methodologies used for graph prediction tasks, our work has some parallels with network time series forecasting, which considers time series of node attributes and forecasts them taking into account the network structure. In this context, a node attribute $x_{i,t}$ is observed for each node $i$ in the network for all time stamps $t$ and the task is to forecast the attribute at future time points \citep{KIM2024971}. In network time series forecasting, the structure of the network is fixed and known. In our case, we forecast the structure of the network at a future time point. To the best of our knowledge, this task has not been studied before. In addition to network time series forecasting, network structure forecasting has some similarities with link prediction. However, our work is different because link prediction considers if a link exists between two nodes $u$ and $v$ where both  nodes $u$ and $v$ are known to exist in the future network. Thus, the nodes in the network at time $t+1$ are known in link prediction. Furthermore, link prediction studies divide links into training and test sets and trains a model on the links in the training set and tests it on the unseen links \citep{Kumar2020}. Link prediction does not forecast the structure of the full network, which is our focus.

Forecasting the structure of dynamic networks have applications in energy, social and disease networks. While traditionally energy networks focused on aggregate measurements, recent advances such as smart energy initiatives need more adaptable techniques and forecasts. For example, behind the meter forecasting and monitoring \citep{behindthemeter1, bauknecht2020behind}  and peer-to-peer trading in energy networks \citep{ALAM20191434} call for newer methods as the network can constantly evolve. Smart homes can act as energy providers coining the term ``prosumers'' in energy network literature.  Not only electric vehicles move locations, but they can act as power sources changing the network structure \citep{Alshahrani2019}.  In social networks, the network structure is dynamic and network growth and evolution are topics of interest \citep{SOBKOWICZ2012470, Wu2017} especially as they can model responses to certain events. Similarly, modeling disease spread in a network by contact \citep{Read2003} and the evolution of infectious diseases \citep{Leskovec2007} consider dynamic network structures.

In this paper, we consider simple graphs, i.e., unweighted, undirected graphs without any node or edge attributes. The proposed method is motivated by Flux Balance Analysis (FBA) \citep{Orth2010}. FBA is a mathematical approach widely used in biochemistry to reconstruct metabolic networks by using linear and mixed-integer programming. By solving an optimization problem maximizing the biomass function subject to a large number of constraints, FBA arrives at the flux solution describing a network of chemical reactions. We adapt FBA to a graph/network forecasting framework by considering the degree of each node instead of the flux of chemical reactions. To forecast the graph at a future time step, we incorporate the degree forecast of each node in the constraints. Thus, the adapted FBA solution gives us a graph satisfying the degree forecasts; this is the graph forecast. Our contributions are summarized as follows: 

\begin{itemize}
    \item We address the challenge of forecasting the evolving structure of dynamic graphs accounting for the emergence of new nodes and edges at future time steps.
    \item We propose a novel approach that adapts flux balance analysis (a method used to reconstruct metabolic networks in biochemistry) enabling  graph forecasting.  
    \item We demonstrate the applicability of our method on both synthetic and real-world datasets.
\end{itemize}




\section{Related Work and Notation}\label{sec:related}

In this section we briefly introduce Flux Balance Analysis (FBA) and introduce the notation.

\subsection{Flux Balance Analysis: a quick introduction}\label{sec:fbaintro}
Flux Balance Analysis (FBA) is a mathematical approach widely used in biochemistry to reconstruct metabolic networks from partial information. A large pool of chemical reactions is used as input to this approach. By mathematically representing chemical reactions using  stoichiometric coefficients, FBA comes up with  a set of constraints that need to be satisfied. The solution space defined by the set of constraints describing the potential network is denoted by the matrix equation $\bm{S}\bm{v} = 0$ along with the inequalities $a_i \leq v_i \leq b_i$ for $i \in \{1, \ldots, q \}$ \citep{Orth2010}, where $\bm{v}$ denotes the vector of fluxes to be determined and $\bm{S}$ denotes the constraint matrix. As $\bm{S}$ is a $p \times q$ matrix with $p < q$, there are multiple solutions to this system of linear equations. FBA selects the optimal solution by maximizing a growth function denoted by $Z = \sum_i  c_i v_i $.  

\subsection{Notation}\label{sec:notation}
We consider a time series of undirected, unweighted,  graphs without loops $\mathcal{G}_t = \left(V_t, E_t\right)$ where $V_t$ and $E_t$ denote the set of vertices and edges respectively. We consider the scenario of growing graphs, where $V_t \subseteq V_{t+1}$ and focus on forecasting $\mathcal{G}_{T+h} = \left( V_{T+h}, E_{T+h} \right)$.  Let $V_t = \{v_1, \ldots, v_{n_t}\}$ denote the set of vertices in graph $\mathcal{G}_t$ with $\vert V_t \vert = n_t$. Let $e_{ij,t}$ denote the edge between nodes $i$ and $j$ at time $t$, where $i$ and $j$ can denote new or old vertices and let $m_t = \vert E_t \vert$ denote the number of edges in $\mathcal{G}_t$. 
Let $d_{i,t}$ denote the degree of vertex $v_i$ in $\mathcal{G}_t$, i.e., in an unweighted graph setting $d_{i,t} = \sum_j e_{ij,t}$. The edge-density of graph $\mathcal{G}_t$ is given by $2m_t/(n_t(n_t-1))$ which we denote by $\dens(t)$. 

We denote the   union of  graphs by $\mathcal{G}_U = \mathcal{G}_1 \bigcup \ldots \bigcup \mathcal{G}_T$ and the set of $K$ most popular nodes (largest degree) in $\mathcal{G}_U$ by $\kappa$. When computing $\mathcal{G}_U$ multiple edges between the same vertices are collapsed to a single edge, i.e., we do not consider multiple edge graphs. With a slight abuse of notation we denote the degree of node $i$ in $\mathcal{G}_U$  by $d_{i,U}$.  We denote all forecasts by a hat $\hat{}$. For example, the forecast number of vertices at time $T+h$ is denoted by $\hat{n}_{T+h}$. Our aim is to obtain a forecast for $\mathcal{G}_{T+h} = \left( V_{T+h}, E_{T+h} \right)$, which we denote by  $\hat{\mathcal{G}}_{T+h} =  \left( \hat{V}_{T+h}, \hat{E}_{T+h} \right)$ containing the  structure of the graph at $T+h$ including  new vertices and new edges.  


\section{Methodology}\label{sec:methods}
\begin{figure}[t]
    \centering
    \includegraphics[scale=0.25]{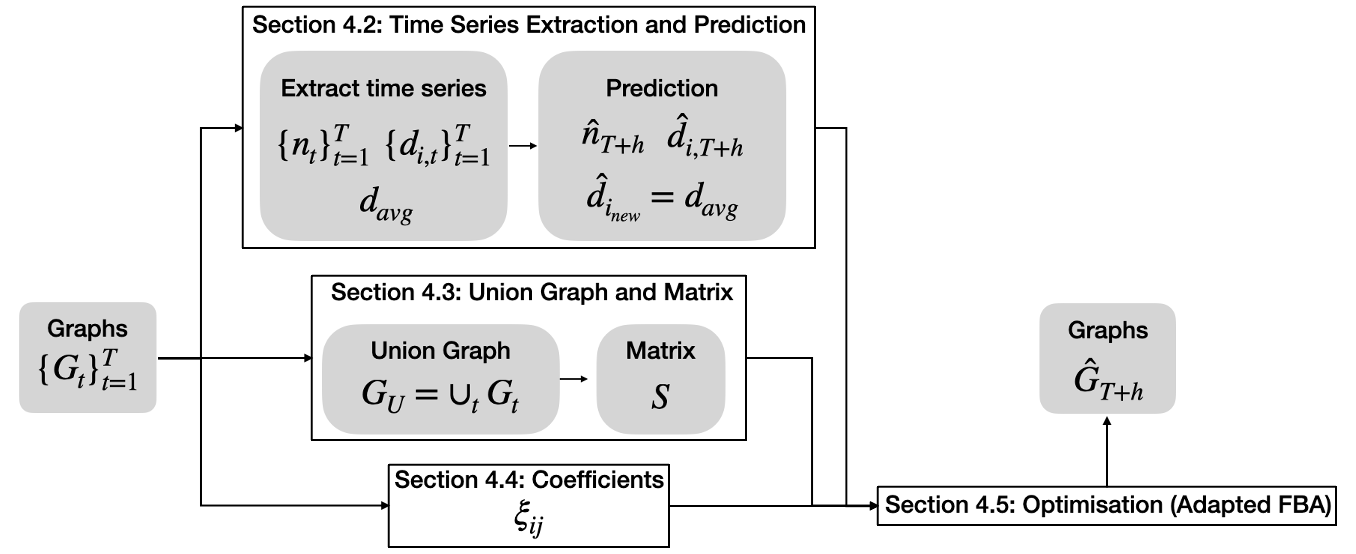}
    \caption{The graph time series $\{\mathcal{G}_t\}_{t=1}^T$ is used to extract multiple time series: $n_t$ is the number of nodes in $\mathcal{G}_t$ and $d_{i,t}$ is the degree of each node $i$ at time $t$.  The forecast $\hat{n}_{T+h}$ is computed using the time series $\{n_t\}_{t=1}^T$ and the forecast of the degree of existing nodes $\hat{d}_{i, T+h}$ is computed using the time series $\{d_{i,t}\}_{t=1}^T$ using ARIMA models. The mean degree of nodes that appear for the first time, $d_{avg}$ is computed. The number of new nodes $n_{new} = \hat{n}_{T+h} - n_T$. Using $\{\mathcal{G}_t\}_{t=1}^T$ the union graph $\mathcal{G}_U$ and matrix $S$ are computed. Coefficients for optimization $\xi_{ij}$ are determined depending on the scheme.  Finally, adapted Flux Balance Analysis is carried out giving $\hat{\mathcal{G}}_{T+h}$ as the output.}
    \label{fig:methodology}
\end{figure}

FBA is generally used to model the flux, which can be thought of as a flow. In our scenario, we are not modelling a flow. As such, we adapt FBA to solve the problem of graph forecasting. Figure \ref{fig:methodology} gives a flowchart of the proposed methodology. 

\subsection{Adapting FBA to graph forecasting}\label{sec:lptographs}
In metabolic networks FBA is used for network reconstruction, not forecasting. We adapt FBA for graph forecasting by using degree forecasts of individual nodes. We consider an optimization problem where the constraints describe the degree of vertices and each variable denotes the presence/absence of an edge. The sum of the edges of a vertex equals its degree. But as we are forecasting the graph at time $T+h$ we do not know the degree of the vertices at time $T+h$. This is not a problem because we can use the forecasts $\hat{d}_{i,T+h}$ instead. Using the forecasts $\hat{d}_{i,T+h}$ forces us to consider the type of constraints we use, i.e., if we employ strict equalities with each equation specifying the degree of a vertex, then it is possible that such a graph cannot exist because the constraints force the solution space to be empty. For example, an undirected, unweighted graph without loops of 3 vertices with degrees 2, 2, 1 cannot exist, but with degrees 2, 1, 1 or 2, 2, 2 can exist.  
Therefore, we need to incorporate some variability that enables the existence of feasible graphs. This can be achieved by using inequalities instead of strict equalities and by specifying reasonable upper bounds for the vertex degree, such as the upper confidence bound of the degree forecast.  These adjustments can ensure a non-empty solution space. Thus, a constraint is of the form,
\begin{equation}\label{eq:bounds1}
    \sum_j \hat{e}_{ij,T+h} \leq f_u\left(\hat{d}_{i,T+h}\right),
\end{equation}
where $f_u$ denotes an upper bound function. Typically we use the upper confidence limit of the degree forecast as $f_u$. 

\subsection{Time series extraction and forecasting}
In our adaptation, FBA uses time series forecasts, the matrix $S$ and coefficients $\xi_{ij}$ as inputs. The first step is to extract certain time series from $\{ \mathcal{G}_{t} \}_{t = 1}^T$ and forecast them at $T+h$.  The degree of each vertex $d_{i,t}$ in $\mathcal{G}_t$  can be considered as a time series $\{d_{i,t}\}_{t=t_{0,i}}^T$ starting from time $t_{0,i}$ when the $i$th vertex appears. Similarly the time series $\{n_t\}_{t=1}^T$ keeps track of the number of vertices in $\mathcal{G}_t$.  We can forecast each time series $\{d_{i,t}\}_{t=t_{0,i}}^T$ and obtain a point forecast and its confidence intervals using standard time series methods. We use ARIMA models with automatic parameter selection \citep{forecasting} to obtain the point forecasts and their upper confidence limits.  Let us call our point forecasts $\hat{d}_{i,T+h}$ and $\hat{n}_{T+h}$ respectively. Let $u$ denote the percentile used to obtain the upper confidence bound of $\hat{d}_{i,T+h}$, which we call $f_u(\hat{d}_{i,T+h})$. While we can use a percentile $\gamma$ to obtain an upper bound of $\hat{n}_{T+h}$, we generally use the point forecast $\hat{n}_{T+h}$ corresponding to $\gamma = 0.5$. If $\hat{n}_{T+h} > n_{T}$ we can add $\hat{n}_{new} = \hat{n}_{T+h} - n_{T}$ new nodes to the graph. 

From the time series of graphs $\left\{ \mathcal{G}_t \right\}_{t=1}^T$ we know the new nodes at each time step and their degree distribution. We take the average degree of new nodes $d_{avg}$ as the forecast degree $\hat{d}_{i,T+h}$ for new, unseen nodes $v_i$. Thus, we have degree forecast for all nodes, old and new.

\subsection{The matrix $S$ and the union graph $\mathcal{G}_U$ }
For a graph with $n$ vertices, we consider a zero-inflated incidence matrix $S$ of size $n \times n(n-1)/2$.  The incidence matrix of a graph has $n$ rows and $m$ columns where $n$ denotes the number of nodes and $m$ the number of edges.  The $ij$th entry of the matrix is set to $1$ if vertex $v_i$ is incident with edge $e_j$. Our version of the incidence matrix is inflated with zero-columns to incorporate edges that are not present in the graph. We obtain $S$ from the adjacency matrix $A_{n \times n}$ as follows: The columns represent the edges obtained from  unrolling the upper triangular part of $A$ row by row, and the rows of $S$ represent the vertices 1 to $n$. If an edge $e_{ij}$ is present, i.e., $e_{ij} = 1$  with $i < j$ then $s_{ik} = s_{jk} = 1$ where $k$ denotes the column corresponding to edge $e_{ij}$, with $k = \left((n-1) + (n-2) + \ldots + (n-i+1)\right) + (j - i) = \frac{1}{2}(i-1)(2n - i ) + ( j -i)$. 

For example suppose $\mathcal{G}$ is a 4-vertex graph. Thus $S$ would be a $4 \times 6$ matrix with columns corresponding to edges  $\left \{e_{12}, e_{13}, e_{14}, e_{23}, e_{24}, e_{34} \right\}$ and rows corresponding to $\{v_1, v_2, v_3, v_4 \}$ respectively. Suppose $\mathcal{G}$  has edges  $e_{12}$, $e_{14}$, $e_{23}$ and $e_{34}$. Thus, we obtain 
\begin{figure}[!h]
	\centering
	\includegraphics[scale=0.3]{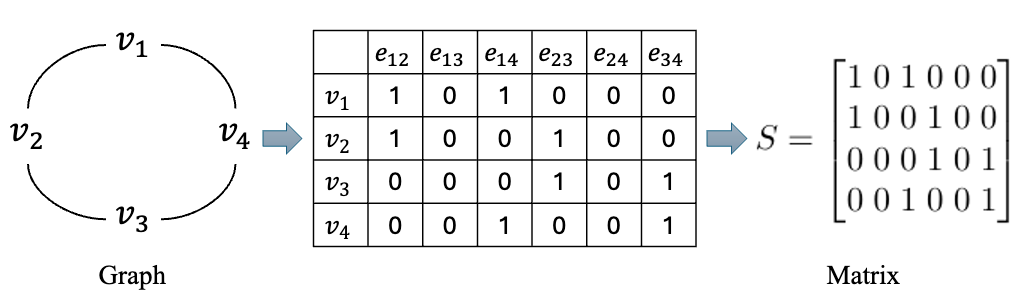}
\vspace{-0.6cm}
\end{figure}

\noindent where for each column corresponding to $e_{ij} = 1$, row $i$ and $j$ are set to 1 and others to zero. 

The matrix $S$ for forecasting the graph at $T+h$ is obtained by considering the union graph from the time series of graphs and  new vertices as per the forecast $\hat{n}_{T+h}$.

\begin{definition}{\textbf{($\bm{S}$ from a graph sequence $ \bm { \{ {\mathcal{G}_t} \}_{t=1}^T}$)}}
For a sequence of graphs $\{ {\mathcal{G}_t} \}_{t=1}^T$ we consider the graph union $\mathcal{G}_U = \mathcal{G}_1 \bigcup \ldots \bigcup \mathcal{G}_T$, and add vertices $\hat{n}_{T+h} - n_T$, if this quantity is positive.  Furthermore, we add edges from new vertices to verticies in $\kappa$ (the $K$ most popular nodes in the union graph $\mathcal{G}_U$) and compute matrix $S$. 

\end{definition}
For a time series of graphs $\{\mathcal{G}_t\}_{t=1}^T$, by obtaining $S$ as above we can denote the constraints in equation \eqref{eq:bounds1} as
\begin{equation}\label{eq:bounds1mat}
    S\bm{u} \leq f_u\left(\bm{d}\right)
\end{equation}
where $\bm{u}$ denotes an $n(n-1)/2$ vector of edges and $\bm{d}$ denotes an $n$ vector of degree forecasts with $n = \hat{n}_{T+h}$. For all new (unseen) vertices $k$ we use the same upper bound $ f_u\left(\hat{d}_{k,T+h}\right) $ computed by taking a mean over the node degree for new nodes from the graph time series $\{ \mathcal{G}_t \}_{t=1}^T$ for $t \leq T$. That is, for each graph $\mathcal{G}_t$ in the time series, we consider the new nodes at time $t$ and take their average degree. For the remainder of this section we use $n$ to denote $\hat{n}_{T+h}$. 




\subsection{Coefficients $\xi_{ij}$}
Of the many graph forecasts possible, we want to select a solution that depicts the most likely graph at time $T+h$. However, the term ``most likely'' can be subjective. Our definition of most likely involves maximizing the function $\sum_{ij} \xi_{ij}\hat{e}_{ij, T+h}$ where $ \xi_{ij}$ denote the coefficients of the optimization function.  The coefficients satisfy $0 \leq \xi_{ij} \leq 1$ with higher values given to edges more likely to be present in $\mathcal{G}_{T+h}$. We state the complete optimization problem below:
\begin{align}\label{eq:optimiseprob}
    \max  \sum_{i,j \in V_{T+h}}  \xi_{ij}\hat{e}_{ij, T+h} \, , \text{with}~~  \bm{0}  \leq   S\bm{u}  \leq f_u\left(\bm{d}\right) \, ,  
 \end{align}
with $\bm{u} = \{u_\ell\}_{\ell = 1}^{n(n+1)/2}$ and $u_\ell \in \{0, 1\}$.
We can consider different formulations for coefficients including the following:
\begin{enumerate}
    \item Uniform coefficients (C1): $\xi_{ij} = 1$ for all $i, j$ \\
    This coefficient function is naive as it considers any edge to be equally likely including edges that have never been present in the past. This function may be useful for instances with limited history or  when we want to completely disregard the history of a network when forecasting edges at time $T+h$. 
    \item Binary coefficients (C2): 
    \begin{equation}\label{eq:binarycoefficients}
        \xi_{ij}= \begin{cases}
        1 & \text{if } e_{ij,t} = 1 \text{ for any } t \leq T  \\
        1 & \text{if } i \text{ or } j \text{ are new vertices} \\
        0 & \text{otherwise}
        \end{cases} 
    \end{equation}
    This scheme assigns $\xi_{ij} = 0$ for existing vertices $i, j$ if $e_{ij, t} = 0$ for all $t \leq T$. For all other cases including edges between new vertices it assigns 1. 
    \item Proportional coefficients (C3): Here we use
        \begin{equation}\label{eq:propcoefficients1}
        \xi_{ij} = \begin{cases}
         \xi_{ij}^{\text{exist}} &  \text{for existing vertices } i \text{ and } j \\
        \xi_{ij}^{\text{new}}  &  \text{ for new vertices } j
        \end{cases} \, . 
    \end{equation}

    For existing vertices $i$ and $j$ we have
    \begin{equation}
        \xi_{ij}^{\text{exist}}  = 
        \frac{ \sum_{t \leq T} \bm{1}_{e_{ij, t} \in \mathcal{G}_t}}{\max_{i,j} \sum_{t \leq T} \bm{1}_{e_{ij, t} \in \mathcal{G}_t} } 
    \end{equation}
    
  where $\bm{1}$ is the indicator function  
  \begin{equation}
      \bm{1}_{e_{ij, t} \in \mathcal{G}_t} = \begin{cases}
      1  \quad \text{if } e_{ij, t} \in \mathcal{G}_t  \\
      0 \quad \text{otherwise}    
      \end{cases} \, . 
  \end{equation}
  The term $\sum_{t \leq T} \bm{1}_{e_{ij, t} \in \mathcal{G}_t}$ counts the  number of times edge $e_{ij}$ is present in the graph time series. Thus, for existing vertices at time $T$, we take the proportion of time points when edge $e_{ij}$ is present.
   For new vertices $j$ connecting to existing vertices $i$ we have
  $$ \xi_{ij}^{\text{new}}  =  Q\left( \frac{d_{i,U}}{\sum_{k \in \kappa} d_{k,U} }\right)  \, , 
  $$
  where $Q$ is the quantile function with respect to the   distribution of $\xi_{ij}^{\text{exist}} $ where both $i$ and $j$ are existing vertices. The term $d_{i, U}$ denotes the degree of node $i$ in the union graph. The $\frac{d_{i,U}}{\sum_{k \in \kappa} d_{k,U} }$th quantile used. For new vertices, we assign a larger weight if it connects to a more popular existing node. Recall that edges are added to $\mathcal{G}_U$ connecting new nodes to nodes in $\kappa$, the $K$ most popular nodes in $\mathcal{G}_U$. The weights for these edges depend on the degree of the node $v_i \in \kappa$. We want the distribution  $\xi_{ij}^{\text{exist}} $ to match with $\xi_{ij}^{\text{new}} $. To achieve this we take quantiles proportional to $d_{i,U}$ from $\xi_{ij}^{\text{exist}}$. 
  
    \item Linear-decay coefficients (C4): 
     \begin{equation}\label{eq:propcoefficients2}
        \xi_{ij} = \begin{cases}
           \frac{ \sum_{t \leq T} t \cdot \bm{1}_{e_{ij, t} \in \mathcal{G}_t}}{\max_{i,j} \sum_{t \leq T} t \cdot \bm{1}_{e_{ij, t} \in \mathcal{G}_t} } &  \text{for existing vertices } i \text{ and } j\\
         Q\left( \frac{d_{i,U}}{\sum_{k \in \kappa} d_{k,U} }\right)  &  \text{if }  j \text{ is a new vertex} 
        \end{cases}
    \end{equation}
    This is a linearly weighted version of C3 coefficients with higher weights given to more recent time points.  Thus, the coefficients linearly decay with time.     
    \item Harmonic-decay coefficients(C5):
    \begin{equation}\label{eq:propcoefficients3}
    \xi_{ij} = \begin{cases}
       \frac{ \sum_{t \leq T} \left(  \frac{1}{T - t + 1} \cdot \bm{1}_{e_{ij, t} \in \mathcal{G}_t}  \right) }{\max_{i,j} \sum_{t \leq T} \left(  \frac{1}{T - t + 1} \cdot \bm{1}_{e_{ij, t} \in \mathcal{G}_t} \right)} &  \text{for existing vertices } i \text{ and } j\\
     Q\left( \frac{d_{i,U}}{\sum_{k \in \kappa} d_{k,U} }\right)   &  \text{if } j \text{ is a new vertex} 
    \end{cases}
    \end{equation}
    This too is a weighted version of C3 with weights for edges of graph $\mathcal{G}_t$ multiplied by $ \frac{1}{T - t + 1}$. Thus, edges in  $\mathcal{G}_T$ would get initially multiplied by 1, and edges in $\mathcal{G}_{T-1}$ would get multiplied by 1/2 and so on.  The coefficients are harmonically decaying in time.     
    \item Last seen graph coefficients (C6): 
    \begin{equation}\label{eq:lastseencoefficients}
    \xi_{ij} = \begin{cases}
       \bm{1}_{e_{ij, T} \in \mathcal{G}_t}  &  \text{for existing vertices } i \text{ and } j\\
     Q\left( \frac{d_{i,U}}{\sum_{k \in \kappa} d_{k,U} }\right)   &  \text{if } j \text{ is a new vertex} 
    \end{cases}
    \end{equation}
    In this scheme we assign 1 to edges existing in the last seen graph $\mathcal{G}_T$ and 0 to other edges in the union graph $\mathcal{G}_U$. For edges from new vertices we use the same quantile function as before. 
\end{enumerate}
The above coefficient schemes can be seen as specific instances of a generalized version given by  
\begin{equation}\label{eq:predictivecoefficients}
        \xi_{ij} = \begin{cases}
        f\left(e_{ij}\right) & \text{ for existing vertices } i, j \\
        \mu_i & \text{ if } v_i \text{ is existing and } v_j \text{ is new } \\
        \end{cases}
\end{equation}
where $f\left(e_{ij}\right)$ is some prediction of $e_{ij}$ and $\mu_i$ depends on existing node  $v_i$. 

Considering the coefficients $\{\xi_{ij}\}_{v_i, v_j \in V_{t+H}}$, the question arises why these alone cannot be used to describe the graph $\mathcal{G}_{T+h}$. We emphasize that $\xi_{ij}$ is computed by considering individual edges $e_{ij}$ or nodes $v_i, v_j$, without taking the network structure into account. Thus, $\xi_{ij}$ is not a good approximation for $e_{ij, T+h}$ and we do not recommend it to be used by itself to construct $\mathcal{G}_{T+h}$. However, $\xi_{ij}$ provides good value as coefficients of the optimization problem described in  equation \eqref{eq:optimiseprob}. 

\subsection{The optimization problem}
We specify two versions of the optimization problem. The first version has the individual degree constraints discussed so far, and the second version has an additional constraint on total edges.  

\subsubsection{Formulation 1 (F1)}
Consider the following optimization problem:
\begin{align}\label{eq:optimiseprob2}
       \max  \sum_{i,j \in V_{T+h}}  \xi_{ij}\hat{e}_{ij, T+h} \, ,  \text{with}~~ \bm{0} \leq   S\bm{u}   \leq f_u\left(\bm{d}\right) \, ,  
 \end{align}
with $\bm{u} = \{u_\ell\}_{\ell = 1}^{n(n+1)/2}$ and $u_\ell \in \{0, 1\}$ where each inequality in \eqref{eq:optimiseprob2} describes a constraint for the degree upper bound of a single vertex $v_i$ given by
$$  \sum_{j} \hat{e}_{ij, T+h} \leq f_u \left( \hat{d}_{i, T+h} \right) \, .
$$

So far each constraint considers the degree of a single node. However, as we are considering upper bounds we may add a lot more edges to the graph. By constraining the total edges using its prediction can help us to get a more realistic graph. We note that this is a global constraint affecting all nodes. Next we use this additional, global constraint in our optimization. 

\subsubsection{Formulation 2 (F2)}
Consider the following optimization problem:
\begin{align*}
    \max & \sum_{i,j \in V_{T+h}}  \xi_{ij}\hat{e}_{ij, T+h} \, ,  \notag \\
 \text{with} \qquad \bm{0}  & \leq   S\bm{u}  \leq f_u\left(\bm{d}\right) \, ,  
 \end{align*}
with $\bm{u} = \{u_\ell\}_{\ell = 1}^{n(n+1)/2}$ and $u_\ell \in \{0, 1\}$ where $S = \{ s_{kl} \}_{(n+1)\times n(n-1)/2}$ with the additional inequality constraining the total edges
\begin{equation} \label{eq:sumofdegree}
    \sum_i\sum_{j} \hat{e}_{ij, T+h} \leq 2 f_u \left( \vert \hat{E}_{T+h} \vert \right) \, ,
\end{equation}
where $\vert \hat{E}_{T+h} \vert$ denotes the predicted number of edges in graph $\mathcal{G}_{T+h}$,  $f_u$ the upper bound function and constant 2 accounts for double counting edges.

It is worth noting the constraints result in a convex space, which guarantees a solution in a real-valued scenario. As we are considering  binary solutions, we explore the cardinality of the solution space to better understand it. 

\begin{theorem}\label{thm:1}
    Let $\mathcal{U} \, , \bm{u} \in \mathcal{U}$ describe the solution space of the optimization problem  described in Formulation 1. Then the cardinality of $\mathcal{U}$, denoted by $\left\vert \mathcal{U} \right\vert$ satisfies the following bounds
    \[ 0 < C_1 \leq  \left\vert \mathcal{U} \right\vert \leq C_2 \]
    where $C_1$ and $C_2$ depend on $n$ and $f_u(\bm{d})$.  
\end{theorem}
The proof is given in the Appendix.

\subsection{$\hat{\mathcal{G}}_{T+h}$ and the forecast distribution of graphs}\label{sec:graphdist}
By solving the optimization problem described in Formulations 1 or 2  we obtain a possible graph at time $T+h$, which we denote by $\hat{\mathcal{G}}_{T+h}$. This graph is a forecast and thus is an element of a forecast distribution. In fact it was obtained by making certain choices on $n$, (used to denote $\hat{n}_{T+h}$) and $f_u(\bm{d})$. 

\begin{figure}[t]
    \begin{center}
    \includegraphics[scale = 0.3]{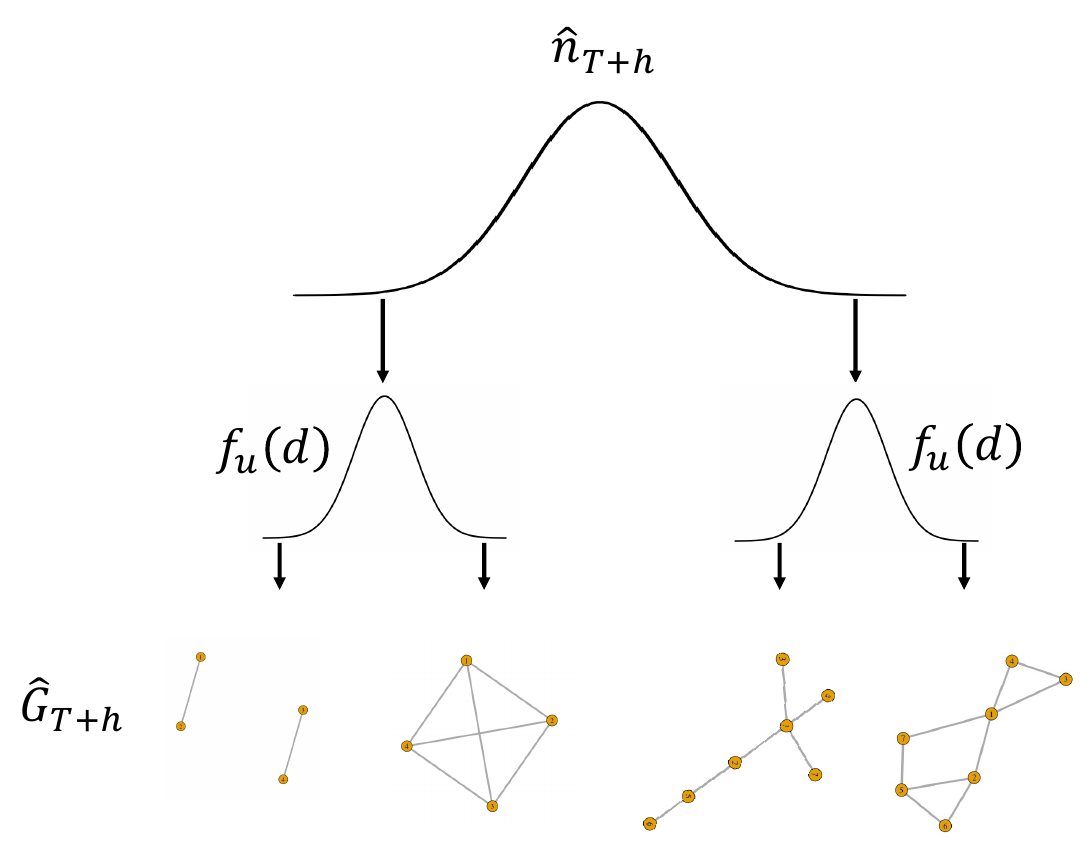}
    \caption{A simple illustration of the forecast graph distribution in terms of $\gamma$ and $u$. The parameter $\gamma$ gives different $\hat{n}_{T+h}$ values. The 2 graphs on the left have 4 nodes each, corresponding to a single $\gamma$ (or $\hat{n}_{T+h}$) value, but has a different number of edges corresponding to different $u$ values. Similarly, the two graphs on the right have 7 nodes corresponding to a higher value of $\gamma$ with the rightmost graph having more edges due to a higher $u$. }
    \label{fig:networkdistribution}
    \end{center}
\end{figure}

Figure \ref{fig:networkdistribution} illustrates the forecast graph distribution in which $\hat{\mathcal{G}}_{T+h}$ is a function of forecast nodes $\hat{n}_{T+h}$ and $f_u(\bm{d})$. If we explore the forecasts for different values of $\hat{n}_{T+h}$ and different upper bounds $u$, we get different graphs. Let $\gamma$ denote the quantile used to obtain $\hat{n}_{T+h}$ and $u$ denote the quantiles in $f_u(\bm{d})$. Then we can think of the graph $\hat{\mathcal{G}}_{T+h}$ as belonging to a graph distribution $\mathfrak{G}$ having parameters $\gamma$ and $u$:
\[
\hat{\mathcal{G}}_{T+h} \sim \mathfrak{G}\left(\gamma, u \right) \, .
\]
While these are explicit parameters, the coefficient scheme used to obtain $\xi_{ij}$, the methods used for forecasting individual time series and their hyperparameters can also be considered as part of a wider parameter pool.


\section{Results}\label{sec:results}
We investigate the performance of the proposed network prediction algorithm on synthetic and real datasets.  For all experiments we consider a time series of graphs $\{ \mathcal{G}_t\}_{t=1}^T$ and forecast graphs at future time steps for $h \in \{ 1, \ldots, 5\} $ using Formulation 2 (F2) with harmonic (C5) and last seen (C6) coefficient schemes. The R package \texttt{netseer} has functionality for both F1 and F2 formulations and other coefficient schemes. For comparison we use the last seen graph $\mathcal{G}_T$. 

\subsubsection{Evaluation Metrics:}\label{sec:evaluation}
We use the following evaluation metrics: (1) Node error  $\frac{ \vert \hat{n}_{T+h} - n_{T+h} \vert }{n_{T+h}}$ gives the absolute error ratio of the number of vertices in the predicted graph compared to the actual graph; (2) Edge error $\frac{ \vert  \hat{m}_{T+h}  -   m_{T+h} \vert }{  m_{T+h} }$ gives the absolute error ratio of the number of edges in the predicted graph compared to the actual graph; (3) Edge density error  ${\vert \hat{\dens}(T+h) - \dens(T+h) \vert}$ gives the absolute error of edge densities between the predicted graph and the actual graph. As the edge density is between 0 and 1 we take the absolute difference, not a ratio. For all 3 evaluation metrics, a lower absolute error is preferred.

\subsection{Synthetic Experiments}\label{sec:synthetic}
For synthetic experiments, we generate a growing sequence of graphs $\left \{\mathcal{G}_t \right \}_{t=1}^T$ using  Preferential Attachment (PA) \citep{barabasi1999}. The PA model considers nodes connecting to more connected nodes with a higher probability. The linear PA model specifies the probability $\Pi(k)$ of a new node connecting to node $i$ with degree $k_i$ to be  
$$ \Pi(k_i) = \frac{k_i}{\sum_i k_i} \, . 
$$
At each time step, the PA model adds a new node with $s$ edges. After $t$ timesteps the network has $n_t = t + s_0$ nodes and $m_t = s_0 + ts$ edges where at time $t = 0$ the network has $s_0$ nodes and edges. 

As the first experiment we consider a sequence of 30 PA networks with $s = 10$, $n_1 = 50$ and $n_{30} = 350$. At each timestep, 5 nodes are added and each node connects to 10 other nodes.  Using the first 25 networks, we forecast the next 5 networks, i.e., we set $T = 25$ and forecast $\hat{\mathcal{G}}_{T+h}$ for $h \in \{1, \ldots, 5\}$. In this experiment $E_t \subset E_{t+1}$ as edges are  added to the existing network. We repeat the experiment 10 times with different seeds to account for randomness. 

The second experiment includes edge deletion. After adding new nodes and new edges as before, at each time step we randomly delete $10$ edges from the network. As such $E_t \not\subset E_{t+1}$. Again, we repeat the computation 10 times to account for randomness. 

\begin{table}[!ht]
\caption{Experiment 1 and 2 Results. Mean and standard deviation in brackets of 10 iterations.}
\label{tab:syntheticexperiments1&2}
\setlength\tabcolsep{3pt}
\begin{center}
    {\small   \begin{tabular}{cccccccc}
    \toprule
    Exp. & Metric & Coef. & $h = 1$ & $h = 2$ & $h = 3$ & $h = 4$ & $h = 5$ \\
    \midrule
    \multirow{9}*{1}   &  \multirow{2}*{Node }     & C5 & 0 (0) & 0 (0) & 0 (0) & 0 (0) & 0 (0) \\
     &  \multirow{2}*{Err. } & C6 & 0 (0) & 0 (0) & 0 (0) & 0 (0) & 0 (0) \\
      &   & LS & 0.0286 (0) & 0.0556 (0) & 0.0811 (0) & 0.1053 (0) & 0.1282 (0) \\
         \cmidrule{2-8}
     & \multirow{2}*{Edge } & C5 & 0 (0) & 0.0077 (0.0014) & 0.0196 (0.0016) & 0.0318 (0.0024) & 0.0435 (0.0025) \\    
     & \multirow{2}*{Err. } & C6 & 0 (0) & 0.004 (7e-04) & 0.0099 (9e-04) & 0.016 (0.0012) & 0.0218 (0.0013) \\
     & & LS & 0.0295 (0) & 0.0573 (0) & 0.0836 (0) & 0.1084 (0) & 0.1319 (0) \\
      \cmidrule{2-8}
    & \multirow{2}*{Dens} & C5 & 0 (0) & 0.0077 (0.0014) & 0.0196 (0.0016) & 0.0318 (0.0024) & 0.0435 (0.0025) \\
     & \multirow{2}*{Err.} & C6 & 0 (0) & 0.004 (7e-04) & 0.0099 (9e-04) & 0.016 (0.0012) & 0.0218 (0.0013) \\ 
     & & LS & 0.0286 (0) & 0.0572 (0) & 0.0858 (0) & 0.1144 (0) & 0.143 (0) \\
    \midrule
    \multirow{9}*{2}   &  \multirow{2}*{Node }     & C5 & 0 (0) & 0 (0) & 0 (0) & 0 (0) & 0 (0) \\
    &  \multirow{2}*{Err.} & C6 & 0 (0) & 0 (0) & 0 (0) & 0 (0) & 0 (0) \\
    & & LS & 0.0286 (0) & 0.0556 (0) & 0.0811 (0) & 0.1053 (0) & 0.1282 (0) \\
    \cmidrule{2-8}
     & \multirow{2}*{Edge } & C5 & 6e-04 (0) & 6e-04 (2e-04) & 0.0087 (0.0028) & 0.0205 (0.0028) & 0.0321 (0.0028) \\
     & \multirow{2}*{Err} & C6 & 0 (0) & 6e-04 (7e-04) & 0.0062 (0.0013) & 0.0119 (0.0013) & 0.0178 (0.0013) \\
     & & LS & 0.0297 (0) & 0.0576 (0) & 0.084 (0) & 0.109 (0) & 0.1326 (0) \\
    \cmidrule{2-8}
    &  \multirow{2}*{Dens.} & C5 & 6e-04 (0) & 6e-04 (2e-04) & 0.0087 (0.0028) & 0.0205 (0.0028) & 0.0321 (0.0028) \\
    &  \multirow{2}*{Err.} & C6 & 0 (0) & 6e-04 (7e-04) & 0.0062 (0.0013) & 0.0119 (0.0013) & 0.0178 (0.0013) \\
    & & LS & 0.0284 (0) & 0.0568 (0) & 0.0853 (0) & 0.1137 (0) & 0.1421 (0) \\
    \bottomrule
    \end{tabular}}
\end{center}
\end{table}

\begin{figure}[t]
    \centering
    \includegraphics[width=0.33\columnwidth]{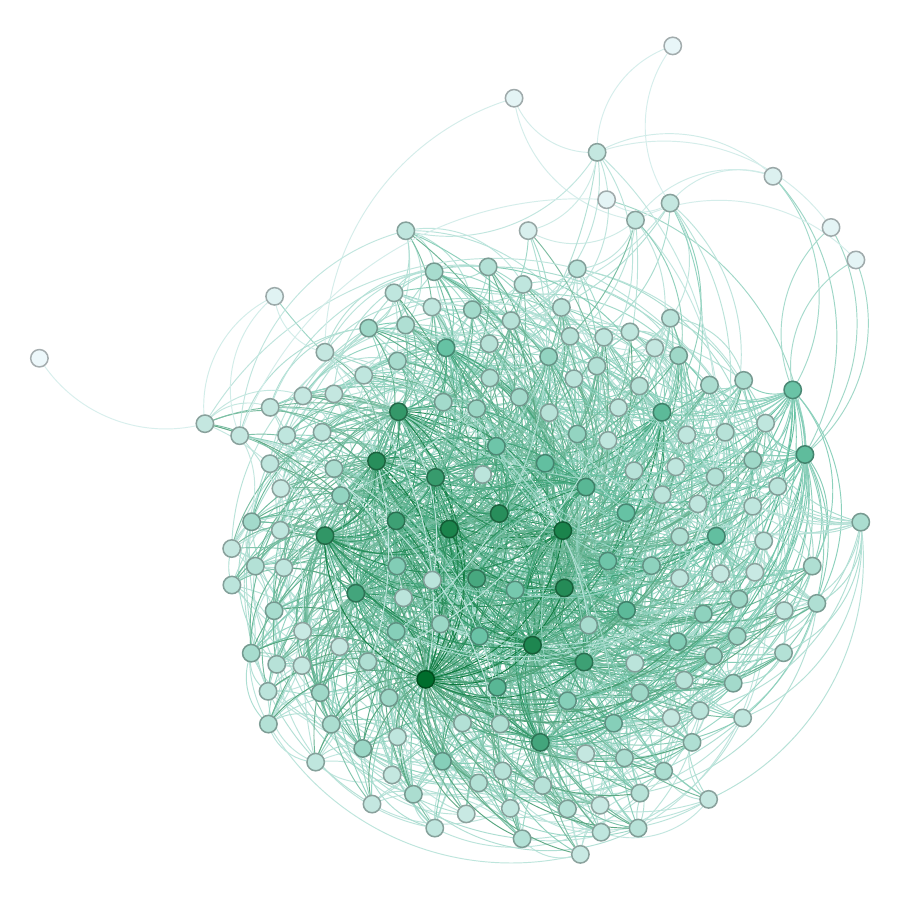}%
    \includegraphics[width=0.33\columnwidth]{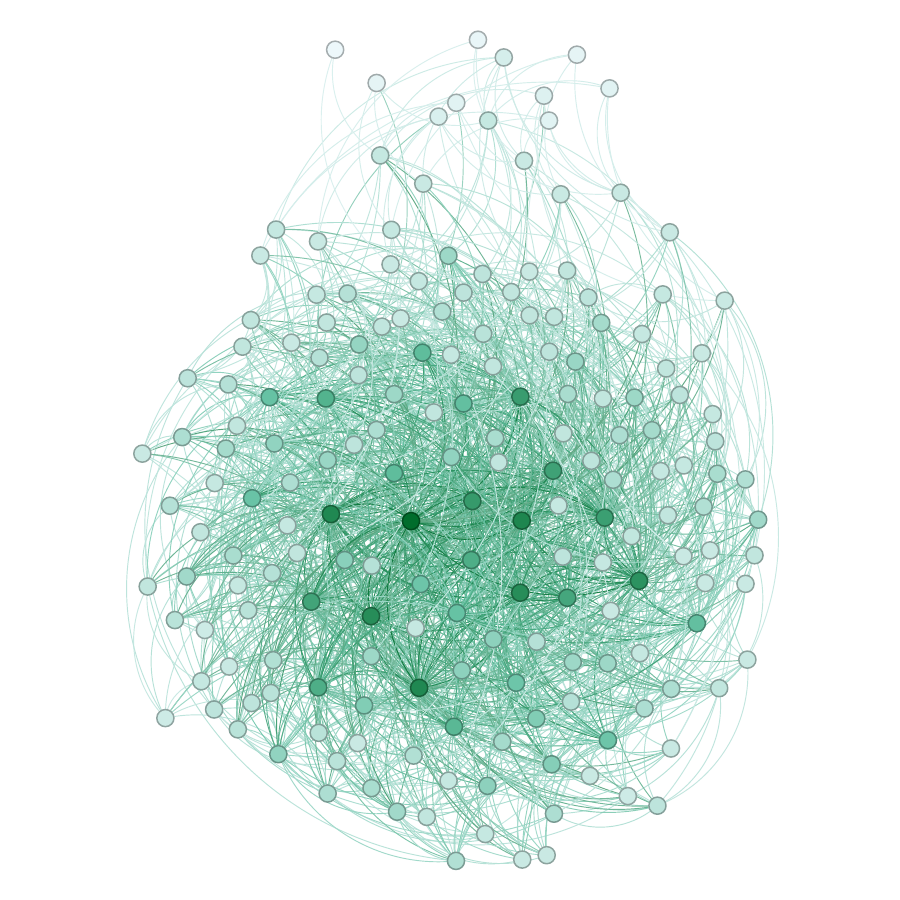}%
    \includegraphics[width=0.33\columnwidth]{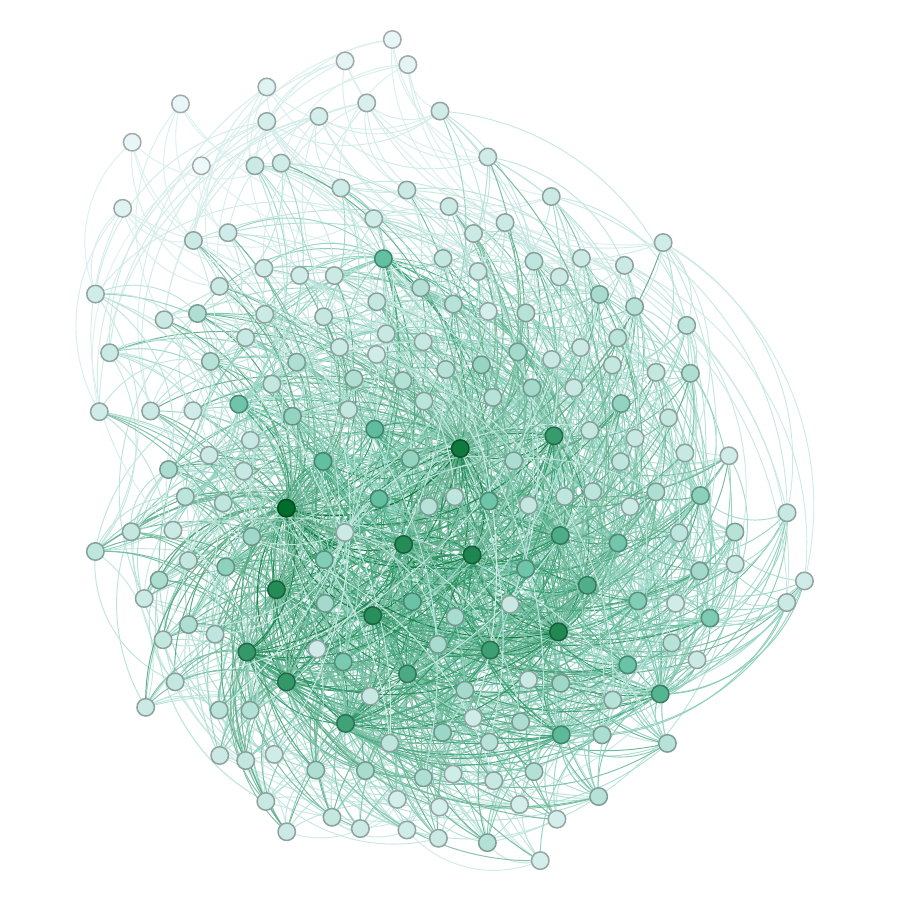}%
    \caption{Forecast synthetic networks using a sequence of 25 networks for $h  \in \{ 1, 3, 5  \}$ illustrating the forecast graphs at time steps 26, 28 and 30.  Node colour depicts the degree of the node, with darker nodes having a higher degree.}%
    \label{fig:networkstructuressynthetic}
\end{figure}

Table \ref{tab:syntheticexperiments1&2} gives the evaluation metrics for the first and second experiments. We see that for both experiments coefficient schemes C5 and C6 achieve better results compared to the last seen network, which is denoted by LS.  The node error of C5 and C6 coefficient schemes is 0 for both experiments. The edge error is quite small for both C5 and C6 compared to LS. Similarly the edge density error for C5 and C6 is much smaller than that of LS.

Figure \ref{fig:networkstructuressynthetic} illustrates the  graph forecasts for synthetic experiment 2 for time steps $h \in \{1, 3, 5\}$. These are graph forecasts at time steps 26, 28 and 30 using $\gamma = 0.5$ and $u = 0.55$.  The node color reflects the degree of the node with darker shades denoting nodes with higher degree.  We see the algorithm has accounted for graph growth. 

\subsection{Experiments using real-world datasets}

\begin{figure}[!ht]
    \centering
    \includegraphics[width=0.8\linewidth]{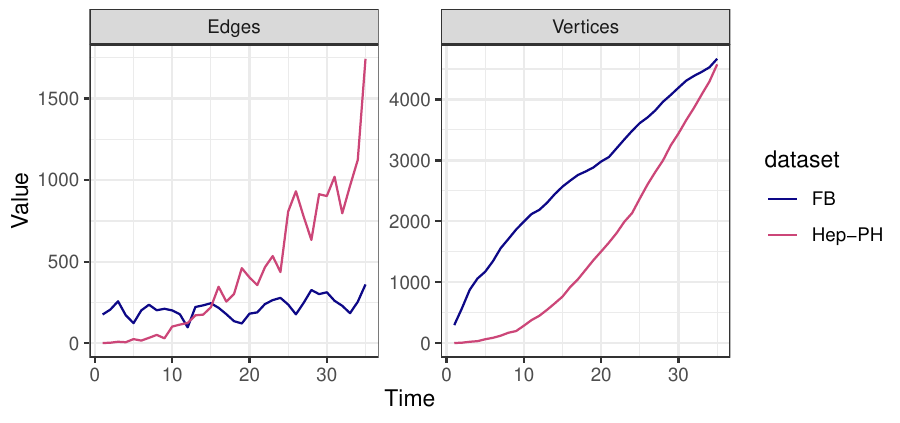}
    \caption{Edges and vertices in each network for realworld datasets.}
    \label{fig:datasetnodesedges}
\end{figure}

We use two realworld datasets in this study:
\begin{enumerate}
   \item \textbf{Facebook (FB)}: This dataset contains users and their links from the Facebook New Orleans networks \citep{viswanath-2009-activity}. Users are depicted as nodes in the network and an edge exists between two users if they are friends.  There are timestamped and non-timestamped edges in this dataset. We used timestamped edges for network forecasting. 
   \item \textbf{ Physics citations (Hep-PH)}: This dataset contains citations of high energy physics papers published on ArXiv \citep{Leskovec2007}. Each paper is denoted as a node and citations between papers are denoted as edges. 
\end{enumerate}
Both FB and Hep-PH datasets are anonymized edge lists of observations of the form $(u, v, t)$. For FB  we consider daily edges to form a network and for Hep-PH we consider monthly edges. For both networks $E_t \not\subset E_{t+1}$ because only edges in each time window are used to construct the network. Figure \ref{fig:datasetnodesedges} shows the number of edges and vertices in each network $\mathcal{G}_t$ for $t \in \{1, \ldots, 35\} $. 

To standardize computations across the two datasets we consider the network time series $\{\mathcal{G}_t \}_{t = 1}^T$ with $T \in \{20, \ldots, 30 \}$ and forecast $\hat{\mathcal{G}}_{T + h}$ for $h \in \{1, \ldots, 5\}$. Forecasts were carried out for all combinations of $T$ and $h$.  We used Formulation 2 (F2) with harmonic decay (C5) and last seen graph coefficients (C6) to forecast the network. For all configurations we used 0.55 as the upper bound $u$ in $f_u(\bm{d})$. As a means of comparison, we used the last seen (LS) graph  $\mathcal{G}_T$ and computed the evaluation metrics for LS.  

Table \ref{tab:realworld} shows the graph forecasting results for FB and Hep-PH datasets. The node, edge and density error for coefficient schemes C5 and C6 are given along with the metrics of the last seen (LS) graph.  We see that for almost all timesteps $h$, C5 and C6 perform better than LS in all metrics. The only exception is the edge error for $h=1$ in Hep-PH, for which LS achieves a lower value than C5 or C6. For both FB and Hep-PH, the node error is much lower for C5 and C6 compared to LS. The edge error too is lower for C5 and C6. All edge density errors are low as these graphs are sparse even though C5 and C6 achieve lower errors. We also see that   several timesteps ahead forecasts have larger errors compared to forecasts with shorter timesteps, which is to be expected. 

\begin{table}[!ht]
\caption{Results for FB and Hep-PH datasets. Coefficient schemes C5 and C6 are considered. The last seen graph (LS) is considered as a means of comparison. Evaluation metrics node, edge and density errors are given with mean and standard deviation in brackets for 10 iterations. }
\label{tab:realworld}
\setlength\tabcolsep{3pt}
\begin{center}
    {\small   \begin{tabular}{cccccccc}
    \toprule
    Dataset & Metric & Coef. & $h = 1$ & $h = 2$ & $h = 3$ & $h = 4$ & $h = 5$ \\
    \midrule
    \multirow{9}*{FB }   &  \multirow{2}*{Node }     & C5 & 0.0064(0.0062) & 0.0126(0.0115) & 0.0195(0.0155) & 0.0275(0.0176) & 0.0315(0.0194) \\
     &  \multirow{2}*{Err. } & C6 & 0.0064(0.0062) &	0.0126(0.0115) & 	0.0195(0.0155) &	0.0275(0.0176) &	 0.0315(0.0194) \\
      &   & LS & 0.0329(0.0076) & 0.0641(0.0128) &	0.0917(0.0188) & 0.1162(0.0244) &	0.1392(0.0271) \\
         \cmidrule{2-8}
     & \multirow{2}*{Edge } & C5 & 0.1379(0.0797) & 0.1514(0.1020) & 0.1826(0.1143) & 0.1686(0.1156) & 0.1763(0.1177) \\    
     & \multirow{2}*{Err. } & C6 & 0.1353(0.0777) & 0.1530(0.1045) &	0.1826(0.1143) & 0.1686(0.1156) & 0.1763(0.1177) \\
     & & LS & 0.1586(0.1047) & 0.2348(0.1689) & 0.3097(0.1805) & 0.2812(0.1750) & 0.2107(0.2049) \\
      \cmidrule{2-8}
    & \multirow{2}*{Dens} & C5 &   4e-06(2e-06) & 5e-06(3e-06) & 5e-06(3e-06) &	5e-06(3e-06) & 6e-06(5e-06) \\
     & \multirow{2}*{Err.} & C6 & 4e-06(1e-06) & 5e-06(3e-06) & 5e-06(3e-06) & 5e-06(3e-06) & 6e-06(5e-06) \\ 
     & & LS & 5e-06(3e-06) &	8e-06(5e-06) & 	1e-05(5e-06) & 9e-06(6e-06) & 8e-06(6e-06) \\
    \midrule
    \multirow{9}*{Hep-PH }   &  \multirow{2}*{Node }     & C5 & 0.0120(0.0083) & 0.0157(0.0156) & 0.0237(0.0188) & 0.0325(0.0258) & 	0.0441(0.0319) \\
    &  \multirow{2}*{Err.} & C6 &  0.0120(0.0083) &	0.0157(0.0156) & 0.0237(0.0188) &0.0325(0.0258) & 0.0441(0.0319) \\
    & & LS & 0.0775(0.0148) & 0.1458(0.0241) & 0.2064(0.0329) &	0.2594(0.0406) &0.3084(0.0473) \\
    \cmidrule{2-8}
     & \multirow{2}*{Edge } & C5 & 0.2250(0.1374) & 0.2139(0.2028) &	0.2314(0.1192) & 0.1588(0.1484) & 0.2380(0.1346) \\
     & \multirow{2}*{Err} & C6 & 0.2250(0.1374) & 0.2139(0.2028) & 0.2314(0.1192) &	0.1588(0.1484) &0.2380(0.1346) \\ 
     & & LS & 0.1973(0.1174) & 0.2364(0.1665) &	0.2489(0.1508) & 0.2361(0.1737) & 0.3113(0.2063) \\
    \cmidrule{2-8}
    &  \multirow{2}*{Dens.} & C5 &  4e-05(3e-05) & 3e-05(3e-05) &	3e-05(2e-05) &	2e-05(2e-05) &	3e-05(2e-05) \\
    &  \multirow{2}*{Err.} & C6 & 4e-05(3e-05) & 3e-05(3e-05) & 3e-05(2e-05) &	2e-05(2e-05) &	3e-05(2e-05) \\ 
    & & LS &  4e-05(3e-05) & 5e-05(4e-05) & 5e-05(4e-05) & 6e-05(4e-05) & 6e-05(4e-05) \\
    \bottomrule
    \end{tabular}}
\end{center}
\end{table}

\section{Conclusion}\label{conclusion}
We have presented a method to forecast graphs at future time points from a time series of graphs. Our forecast can include new nodes and edges. The proposed method uses time series methods to forecast the node degree and linear programming techniques from flux balance analysis to obtain future graphs. The R package \texttt{netseer}\citep{netseerR} includes the methodology. We investigated the utility of our method on synthetic and real datasets. This work focused on unweighted graphs. Future research avenues include extending this methodology to weighted and directed graphs.

\bibliographystyle{agsm}
\bibliography{references}

\pagebreak
\appendix
\section*{Appendix: Proof of Theorem \ref{thm:1}} \label{sec:Appendix1}
\textit{
    Let $\mathcal{U} \, , \bm{u} \in \mathcal{U}$ describe the solution space of the optimization problem  described in Formulation 1. Then the cardinality of $\mathcal{U}$, denoted by $\left\vert \mathcal{U} \right\vert$ satisfies the following bounds
    \[ 0 < C_1 \leq  \left\vert \mathcal{U} \right\vert \leq C_2 \]
    where $C_1$ and $C_2$ depend on $n$ and $f_u(\bm{d})$.  
} 
\begin{proof}
    For the optimization problem in Formulation 1 we have 
    \begin{align}
        S\bm{u} & \leq f_u\left(\bm{d} \right)  \label{eq:1}\\
        \text{giving us }  \bm{1}^TS\bm{u} & \leq \bm{1}^T f_u\left(\bm{d}   \right) = \sum_i f_u(\hat{d}_i) \label{eq:2}
    \end{align}
    where equation \eqref{eq:2} denotes a single constraint comprising the addition of the individual constraints in equation \eqref{eq:1}. Suppose $\bm{u}_j$ satisfies the individual constraints in equation \eqref{eq:1}. Then  $\bm{u}_j$ satisfies the summed constraint in equation \eqref{eq:2}. Thus $\mathcal{U} \subset \mathcal{U}_{\text{sum}}$ where $\mathcal{U}_{\text{sum}}$ denotes the solution space of the summed constraint. 
    
    We consider $\bm{u}_j \in \mathcal{U}_{\text{sum}}$ as a binary string. Thus, the number of binary strings in $\mathcal{U}_{\text{sum}}$ is an upper bound for $\left\vert  \mathcal{U} \right\vert$.  As $S$ is a zero-inflated incidence matrix, $\sum_{k} s_{kl} = \{0, 2\}$ for all $k$. As a result we have
    $$ \bm{1}^T S = 2  \bm{B}^T
    $$
    where $\bm{1}^T$ is a $1 \times n$ row vector and $\bm{B}^T$ is an $1 \times n(n-1)/2 $ binary row vector (containing ones and zeros).  Substituting this in equation $\eqref{eq:2}$ we get
    \begin{equation}\label{eq:btransposesum}
        2 \bm{B}^T \bm{u} \leq \sum_i f_u(\hat{d}_i) \iff \bm{B}^T \bm{u} \leq \frac{1}{2} \sum_i f_u(\hat{d}_i)
    \end{equation}

    where the hyperplane describing the constraint now has coefficients equal to 1. The number of ones in $\bm{B}$ is equal to the number of edges in the union graph $\mathcal{G}_U$. Let $c_3 = \lfloor \frac{1}{2} \sum_i f_u(\hat{d}_i)  \rfloor$ and $m$ denote the number of ones in $\bm{B}$, which is the length of the binary string we consider. Then the number of binary strings with $k$ ones is ${m \choose k}$. To satisfy the constraint in equation \eqref{eq:2} we can have up to $c_3$ ones giving us a total of
    \begin{equation}\label{eq:upperbound}
        {m \choose 1} + {m \choose 2} + \cdots + {m \choose c_3} = C_2
    \end{equation}

    possible binary sequences comprising the upper bound.

    For the lower bound we consider the order statistics of $f_u(\bm{d})$ in reverse (decreasing) order. Let $f_u(\bm{d})_{(k)}$ be the $k$th reverse order statistic of  $f_u(\bm{d})$ and let $\bm{v}_{(k)}$ be the corresponding vertex.  Let $\eta_i$ be the number of inequalities (corresponding to vertices) in equation \eqref{eq:2} with $f_u(d_k) = i-1$. Thus, $\eta_1$ gives the number of vertices with an upper bound degree 0.  To give some intuition, $\eta_i$ focuses on low  upper bound degree vertices.  Let $k$ be the largest integer that satisfies the following  inequalities:
    \begin{align}
        \eta_k + k + f_u(\bm{d})_{(k)} & \leq n \label{assump1} \\
        f_u(\bm{d})_{(1)} + f_u(\bm{d})_{(2)} + \cdots + f_u(\bm{d})_{(k)} & \leq \frac{1}{2}\sum_i f_u(d_i) \label{assump2} 
    \end{align}
    The first inequality is about selecting vertices to draw edges. When $k = 1$, we consider the vertex $\bm{v}_{(1)}$. There are $n-1$ vertices  available to connect with  $\bm{v}_{(1)}$ if the upper bound of all vertices are strictly positive. But if certain vertices have a zero upper bound, we need to remove those vertices, when considering edges with $\bm{v}_{(1)}$, i.e., then we consider $n-1  - \eta_1$ vertices. The term $f_u(\bm{d})_{(k)} $ denotes the maximum number of edges we can draw as we will see later. 

    The second inequality specifies the number of inequalities $k$ to consider to obtain a lower bound of $|\mathcal{U}|$. As  sum of the degrees is twice the sum of edges for non-weighted graphs, we don't need to consider all the vertices specified by the inequalities. Rather, we only need to focus on the top $\bm{v}_k$ vertices with   upper bounds $\sum_i f_u(\bm{d})_{(i)}$ adding up to the total number of edges. We note that both the upper and lower bounds of $|\mathcal{U}|$ are not strict upper and lower bounds. 
    
    Let us first consider the case when these 2 inequalities are satisfied. We note that the assumptions detailed by the inequalities are reasonable and satisfied for at least $k = 1$ because  $n$ is generally much larger than $f_u(\bm{d})_{(1)}$, the upper bound degree of $\bm{v}_{(1)}$, with $\eta_1$ small, satisfying inequality \eqref{assump1} and the largest upper bound degree $f_u(\bm{d})_{(1)}$ is generally less than all the edges a graph can have or its upper bound $\frac{1}{2}\sum_i f_u(d_i)$. 
    
     We start with $\bm{v}_{(1)}$ the vertex corresponding to $f_u(\bm{d})_{(1)}$, i.e., the vertex with the highest upper bound degree. This vertex can have a maximum number of $f_u(\bm{d})_{(1)}$ edges. First we need to select another vertex to draw an edge from $\bm{v}_{(1)}$. There are $n - 1 - \eta_1$ such vertices because the graph has $n$ vertices and  $u_1$ vertices with upper bound degree 0. An edge can be made from  $\bm{v}_{(1)}$ to any of these vertices. Thus, the number of graphs with 1 edge from $\bm{v}_{(1)}$ are  ${{n - 1 - \eta_1} \choose 1 }$.  To draw 2 edges from $\bm{v}_{(1)}$ we need to select 2 vertices from $n - 1 - \eta_1$. We note that it is $(n - 1 - \eta_1)$ and not $(n - 1 - \eta_2)$ when we are using vertex  $\bm{v}_{(1)}$ because we are counting graphs with 2 edges separately from graphs with 1 edge. The number of graphs with 2 edges from  $\bm{v}_{(1)}$ are ${{n - 1 - \eta_1} \choose 2 }$. Therefore, the number of graphs having upto a maximum of $f_u(\bm{d})_{(1)}$ edges for $\bm{v}_{(1)}$ is
     \begin{equation}\label{eq:g1}
         g_1 = {{n - 1 - \eta_1} \choose 1 } + {{n - 1 - \eta_1} \choose 2 } + \cdots +  {{n - 1 - \eta_1} \choose {f_u(\bm{d})_{(1)}} } \, . 
     \end{equation}

   We note that each of these graphs not only satisfy the largest constraint (with $f_u(\bm{d})_{(1)}$ upper bound degree), but they satisfy the other constraints as well because they draw edges with vertices having degree $ \geq 1$. 
   Having considered the vertex with the largest upper bound degree, we focus on the second largest upper bound degree $f_u(\bm{d})_{(2)}$. When constructing edges from  $\bm{v}_{(2)}$ we only consider vertices other than $\bm{v}_{(1)}$ and vertices that have $f_u(d_i) > 1$ (there are $\eta_2$ such vertices) because we are interested in a lower bound. If we were counting all the graphs, then we need to consider the case when $\bm{v}_{(1)}$  and $\bm{v}_{(2)}$ have an edge and when they don't and compute all possible combinations. As we're interested in a lower bound we simply focus on the other vertices. We have $n - 2 -\eta_2$ such vertices to choose from to draw an edge from $\bm{v}_{(2)}$.  The degree of $\bm{v}_{(2)}$ is bounded by $f_u(\bm{d})_{(2)}$. But we do not know if there is an edge between $\bm{v}_{(1)}$ and $\bm{v}_{(2)}$. To get a lower bound we suppose there is an edge and only allocate up to  $f_u(\bm{d})_{(2)} - 1$ edges from $\bm{v}_{(2)}$. 
   
   The number of graphs satisfying the second largest constraint are
    \begin{equation}\label{eq:g2}
        g_2 \geq {{n - 2 - \eta_2} \choose 1 } + {{n - 2 - \eta_2} \choose 2 } + \cdots +  {{n - 2 - \eta_2} \choose {f_u(\bm{d})_{(2)} - 1} } \, . 
    \end{equation}
    Again, each of these graphs satisfy the other constraints as well. 
    Similarly, the number of graphs satisfying the $k$th largest constraint satisfy
    \begin{equation}\label{eq:g3}
        g_k \geq {{n - k - \eta_k} \choose 1 } + {{n - k - \eta_k} \choose 2 } + \cdots +  {{n - k - \eta_k} \choose {f_u(\bm{d})_{(k)} - k + 1 } } \, . 
    \end{equation}

    For any vertex $\bm{v}_{(i)}$ if $f_u(\bm{d})_{(i)} - i + 1 \leq 0$ then we can stop at $\ell$ such that $f_u(\bm{d})_{(i)} - \ell + 1  = 1$, i.e., stop early in inequality \eqref{eq:g3} without computing all the terms. 
    
    Next we consider the union of graphs satisfying equations \eqref{eq:g1}, \eqref{eq:g2} and \eqref{eq:g3}. The union of any graph specified in equation \eqref{eq:g1} with a graph in equation \eqref{eq:g2}  still satisfies all the constraints because graphs in \eqref{eq:g2} do not draw edges to $\bm{v}_{(1)}$ and so on. Thus, the the graphs satisfying all the constraints is the union of all different combinations of graphs described in each equation/inequality. Therefore we obtain the following lower bound:
    \begin{equation}\label{eq:lowerbound}
         C_1 = g_1 g_2 \ldots g_k
    \end{equation}
     Suppose the assumptions detailed in inequalities \eqref{assump1} and \eqref{assump2} are not satisfied.  Inequality \eqref{assump2} not being satisfied has minimal impact as we will illustrate now. Suppose 
     $$ f_u(\bm{d})_{(1)} \geq \frac{1}{2}\sum_i f_u(d_i) \, . 
     $$
     Then we can compute $g_1$ graphs as detailed in \eqref{eq:g1} and let $C_1 = g_1$. 
     
     
     The assumption in inequality \eqref{assump1} has a more serious effect if it is not met. Suppose $n \leq \eta_1 + 1$, which results in inequality \eqref{assump1} not being met in the most radical way. Then, we have $\eta_1 \geq n -1 $ but as there are only $n$ vertices this implies $\eta_1 = n - 1$ or $\eta_1 = n$ giving us the extreme situation forcing $f_u(\bm{d}) = \bm{0}$ when $\eta_1 = n$. This situation forces there to be no edges giving us the edgeless graph consisting of vertices alone. Thus in this case
     $$ C_1 = 1 \, .
    $$
    Let us consider another situation where  $f_u(\bm{d}) = \bm{1}$. While noting it is unusual to have all the upper bound degrees to be 1, we can still find multiple graphs satisfying this constraint. The first edge can be selected from ${n \choose 2}$  pairs of vertices. The next edge can be selected from ${{n-2} \choose 2}$ pairs of vertices and so on. This results in $\lfloor n /2 \rfloor$ edges, which can be ordered in ${\lfloor n/2 \rfloor}!$ ways. Thus, we would have
    $$ C_1 = \frac{{n \choose 2}{{n-2} \choose 2}\cdots{{q} \choose 2}}{{\lfloor \frac{n}{2} \rfloor }!} \, ,
    $$
    where $q = 2$ or $q = 3$ depending on the graph giving us $C_1$ graphs as the lower bound for this set of constraints. For real graphs it is extremely unlikely to obtain $f_u(\bm{d}) = \bm{1}$ or $f_u(\bm{d}) = \bm{0}$. Thus, mostly we would have a lower bound expressed by equation \eqref{eq:lowerbound}.
\end{proof}

\end{document}